\documentclass[tablecaption=bottom,wcp]{jmlr}% journal article
\jmlrproceedings{arxiv}{arxiv submission}

\usepackage{booktabs}
\usepackage[load-configurations=version-1]{siunitx} % newer version
\usepackage{lineno}
\usepackage{listings}
\usepackage{hyperref}
\usepackage{url}
\usepackage{sidecap}
\usepackage[utf8x]{inputenc}
\usepackage[T1]{fontenc}

\usepackage{amsmath} 
\usepackage{graphicx}
\usepackage{algorithm}
\usepackage[noend]{algpseudocode}
\usepackage{booktabs} % professional-quality tables
\usepackage{graphicx}
\usepackage{subcaption}

\usepackage{titlesec}
\usepackage{xspace}

\newcommand{\ie}{i.e.,\xspace}

\newcommand{\Xf}{\textbf{X}}

\newcommand{\um}[1]{$\textbf{UM}_#1$}
\newcommand{\umf}[1]{$\textbf{UM-Flex}_#1$}
\def\XX{\mathbf{X}}

\def\UM{\mathrm{UM}}

\theorembodyfont{\upshape}
\theoremheaderfont{\scshape}
\theorempostheader{:}
\theoremsep{\newline}

\title[Universal Marginaliser for Deep Amortised Inference for Probabilistic Programs]{Universal Marginaliser for Deep Amortised Inference for Probabilistic Programs}

\author{Robert Walecki, Kostis Gourgoulias, Adam Baker, Chris Hart, Chris Lucas, Max Zwiessele, Albert Buchard, Maria Lomeli, Yura Perov, Saurabh Johri\\
  \addr Babylon Health, London, United Kingdom}
\begin{document}

\maketitle

\begin{abstract}
        Probabilistic programming languages (PPLs) are powerful modelling tools which allow to formalise our knowledge about the world and reason about its inherent uncertainty. 
        Inference methods used in PPL can be computationally costly due to significant time burden and/or storage requirements; or they can lack theoretical guarantees of convergence and accuracy when applied to large scale graphical models. 
       To this end, we present the Universal Marginaliser (UM), a novel method for amortised inference, in PPL.
        We show how combining samples drawn from the original probabilistic program prior with an appropriate augmentation method allows us to train one neural network to approximate any of the corresponding conditional marginal distributions, with any separation into latent and observed variables, and thus amortise the cost of inference.
        Finally, we benchmark the method on multiple probabilistic programs, in Pyro, with different model structure.
\end{abstract}

\section{Introduction}
\vspace{-0.15cm}
By encoding our knowledge of the world in a very expressive statistical formalism, PPLs allow for complex reasoning and decision making under uncertainty. 
% Fear this sentence is too similar to previous work
They have been successfully applied to problems in a wide range of real-life applications including information technology, engineering, systems biology and medicine, among others.
A Bayesian Network (BN) is a particular example of a probabilistic program where nodes and edges are used to define a distribution $P(X, Y)$. Here, $X$ are the latent variables and $Y$ are the observations. 
Ancestral sampling can be used to draw samples from this distribution. 
However, computing the posterior distribution $P(X\ |\ Y)$ is computationally expensive.
If we increase the model complexity, then the cost of inference will increase accordingly, limiting the feasibility of available algorithms. 
Some approximate inference methods are: variational inference~\citep{wainwright2008graphical} and Monte Carlo methods such as importance sampling~\citep{neal2001annealed}. Variational inference methods can be fast but do not target the true posterior. Monte Carlo inference is consistent, but can be computationally expensive. 
Importance sampling methods~\citep{cheng2000ais,neal2001annealed} are efficient MCMC scheme which converge asymptotically to the global optimum. The caveat is that constructing good importance sampling proposals for large programs is hard and either requires expert knowledge~\citep{shwe1991empirical} or is restricted to Bayesian networks with binary nodes \citep{Douetal17}. 
In this work we present an amortised inference technique, the Flexible Universal Marginaliser (\umf{*}), which allows us to ``reuse inferences so as to answer a variety of related queries''~\citep{gershman2014amortized}. More importantly, the model is capable to automatically adapt to different output and input shapes and types (real-valued and categorical).

Amortised inference has been popular for Sequential Monte Carlo and has been used to learn in advance either parameters~\citep{gu2015neural,perov2015data,PaiWoo16,perov2016applications,le2016inference} or a discriminative model ~\citep{Morris:2001:RNA:2074022.2074068, paige2016inference,le2017using,germain2015made}. 
More recently, \cite{ritchie2016deep} applied deep amortised inference to learn network parameters and later perform approximate inference on a Probabilistic Graphical Model (PGM).
Such models either follow the control flow of a predefined sequential procedure, or are restricted to a predefined set of observed nodes. 
Similarly, to the recently introduced Transformer Network architecture \citep{vaswani2017attention}, which use a masking scheme to learn to predict the next word in a sentence, our model relies on masking to learn to approximate the marginals of all possible queries. In contrast, rather than using a fixed mask over nodes, our masking function is probabilistic. Other related methods, such as variational auto encoders \cite{ivanov2018variational} or neural networks \cite{belghazi2019learning} have also been proposed to perform inference given conditional on a subset of variables. However, they are developed mainly for image tasks like inpainting and denoising where they treats each output (pixel) in the same fashion.
Our proposed model is able to learn from the prior samples from a generative model written as a probabilistic program with a bounded number of random choices without any separation into hidden and observed variables beforehand. Furthermore, the outputs can be in categorical and continuous form (or mixed). This allows us to use the same trained discriminative model to approximate any possible posterior $P(X\ |\ Y)$ with any possible separation of variables into latent variables $X$ and observed variables $Y$. This property of one discriminative model being able to amortise inference for any possible separation of nodes into unobserved and observed ones is especially important for complex models of the world and for modelling behaviour of AI agents.
To achieve such property, we use the Universal Marginaliser (UM) \citep{Douetal17,walecki2018universal}, an amortised inference-based method for efficient computation of conditional posterior probabilities in probabilistic programs.

\section{Universal Marginaliser}
\label{sec:um}
\vspace{-0.15cm}
% Cite our previous paper
The Universal Marginaliser (UM) is based on a feed-forward neural network, used to perform fast, single-pass approximate inference on probabilistic programs at any scale.
In this section we introduce the notation and discuss the UM building and training algorithm.\\
\textbf{Notation:}
A probabilistic program can be defined by a probability distribution $P$ over sequences of executions on random variables $\Xf=\{X_1,\ldots, X_N\}$. 
The random variables are divided into two disjoint sets, $Y \subset \Xf$ the set of observations within the program, and $X \subset \Xf\setminus Y$ the set of latent states.
We utilise a neural network to learn an approximation to the values of the conditional posteriors $P(X_i\ |\ Y)$ for each variable $X_i \in \XX$ given an instantiation $Y$ of observations.
For a set of variables $X_i$ with $i \in {1, ..., N}$, the desired neural network maps the vectorised representation of $Y$ to the values $(p_1,\ldots, p_N) = \UM(Y) \approx P(X_i | Y)$.
This NN is used as a function approximator, hence can approximate any posterior marginal distribution given an arbitrary set of observations $Y$. For this reason, such a discriminative model is called the Universal Marginaliser (UM). 
Once the weights of the NN are optimised, it can be used as an approximation for the marginals of hidden variables $X$ given the observations $Y$. It also can be used to compute the hidden variable proposal for each $X_i$ sequentially given all previous $X_1, ..., X_{i-1}$ and observations (i.e. using the chain rule for joint distribution calculation and ancestral sampling algorithm \cite[Algorithm 12.2]{koller2009probabilistic}.
\vspace{-0.2cm}
\subsection{UM architecture}
\vspace{-0.1cm}
In order to train the UM with minimum hyperparameter tuning efforts, we design the UM architecture automatically to the specificities of the target probabilistic program. This is done based on rules that are considered as good practise when building NN.
For example, we deploy categorical cross-entropy (mean square error) loss for nodes with categorical (continuous) states. The number of inner layers, dropout probability and the type of activation function was also selected based on the type of program.
Furthermore, we use the ADAM optimisation method with an initial learning rate of 0.001 and a learning rate decay of $10^{-4}$.
The two model parameters to be set by the user or found by hyperparameter optimisation are $h$, the number of hidden layers and $s$, the number of hidden nodes per layer. 
We suggest to use a deeper and more complex network for larger probabilistic programs.
The UM framework is implemented in the Pyro PPL~\citep{Bingham2018} and the deep learning platform, in PyTorch. 

\subsection{UM training algorithm}
Due to regions of low probability and the combinatorial explosion of the possible queries in complex models, it is often impossible to sample efficiently all possible queries from the generative model. However, the dynamic masking procedure used by the UM generalises well across all queries, learning from smaller batches sampled directly from the probabilistic program. 
This improves memory efficiency during training and ensures that the network receives a large variety of observations, accounting for low probability regions in $P$. The UM is then trained in three steps as follows:
\\
\noindent {\bf A) Sample from program.}
For each iteration, we sample a batch of observations from the prior of the program and use it for training (see appendix Fig.~\ref{example_pp} for an example of such a program).
\\
\noindent {\bf B) Masking.}
In order for the network to approximate the marginal posteriors at test time, and be able to do so for any input observations, each sample $S_i$ is prepared by masking. The network will receive as input a vector where a subset of the nodes initially observed are replaced by the priors. 
This augmentation can be deterministic, \ie always replace specific nodes, or probabilistic. 
We use a constantly changing probabilistic method for masking.
This is achieved by randomly masking $i$ nodes where $i$ is a random number, sampled from a uniform distribution between 0 and $N$. This number changes with every iteration and so does the total number of masked nodes.
\\
\noindent {\bf C) Training multi-output NN.}
We train the NN by minimising multiple losses, where each loss is specifically designed for each of the random variables in the probabilistic program. We use categorical cross-entropy loss for categorical values and mean square error for nodes with continuous values.
We also use a different optimiser for each output and minimise the losses independently. This ensures that the global learning rates are also updated specifically for all random variables. An example of an output of a probabilistic program and the corresponding network is depicted in the appendix (\ref{example_pp}).

\subsection{Experiments}
\begin{figure}[h]
    \subfigure[]{
        \centering
        \includegraphics[width=20mm]{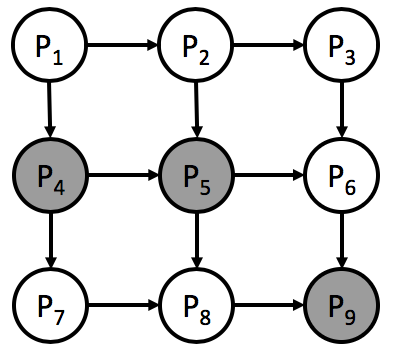}
    }%
    \hfill
    \subfigure[]{
        \centering
        \includegraphics[width=20mm]{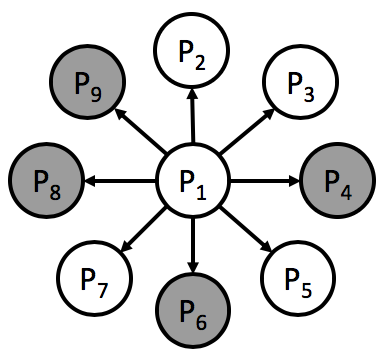}
    }%
    \hfill
    \subfigure[]{
        \centering
        \includegraphics[width=40mm]{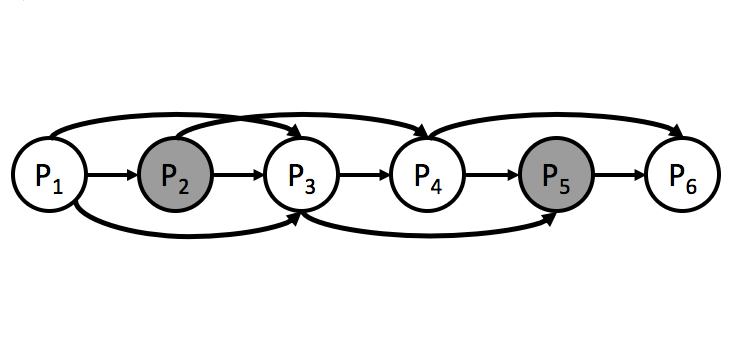}
    }
    \hfill
    \caption{Examples for the different types of graphs. Grid9 (a): grid graph with 9 nodes. Star9 (b): star graph with 9 nodes and Chain6: (c) chain-like structured graph wit 6 nodes.}
    \label{fig:graphs}
\end{figure}

We compared two types of training methods with three different network architectures and eight different probabilistic programs (see Fig.~\ref{fig:graphs}). 
We first used the standard universal marginaliser in form of a neural network, where the losses of all outputs are summed and jointly minimised. We refer to this method as \um{s}, where $s$ indicates the size of the network.
We compared this method with the proposed \textbf{flexible} universal marginaliser. The \textbf{flexible} design allows the method to automatically build the network architecture for different types of probabilistic programs. Furthermore, a different parameter optimiser and loss function was used for each type of outputs. We refer to this method as \umf{s}. 
The architectures of \um{1} and \umf{1} are identical. The networks have 2 hidden layers with 10 nodes each. \um{2} and \umf{2} have 4 hidden layers with 35 nodes each and \um{3} and \umf{3} have 8 hidden layers with 100 nodes. 
The quality of the predicted posteriors was measured by using a test set consisting of 100 sets of observations via importance sampling with one million samples.
Tab.~\ref{tab:results} shows the performance in terms of correlation of various neural networks for marginalisation. Higher correlations indicate a better NN architecture.

\begin{table}[ht]
\scriptsize
        \centering
        \begin{tabular}{l|llllllll}
                & Chain4 & Chain16 & Chain32 & Grid9 & Grid16 & Star4 & Star8 & Star32     \\
                \hline
                \um{1} & 0.903  & 0.875   & 0.698    & 0.877 & 0.926  & 0.914  & 0.822  & 0.667   \\
                \um{2} & 0.932  & 0.852   & 0.795    & 0.824 & 0.904  & 0.920   & 0.821  & 0.804   \\
                \um{3} & 0.927  & 0.837   & 0.631    & 0.843 & 0.919  & 0.900    & 0.756  & 0.783   \\
                \umf{1} &\textbf{0.945}  & 0.859   & 0.703    & 0.875 & 0.928  & 0.919  & 0.907  & 0.697   \\
                \umf{2} & 0.935  & \textbf{0.890 }   & \textbf{0.823}    & 0.889 & \textbf{0.958}  & 0.919  & \textbf{0.908}  & \textbf{0.811}   \\
                \umf{3} & 0.913  & 0.846   & 0.609    & \textbf{0.923} & 0.922  & \textbf{0.933}  & 0.882  & 0.789
        \end{tabular}
        \caption{Results in terms of correlation between conditional posteriors and UM predictions.}
        \label{tab:results}
\end{table}

\subsection{Discussion}
\vspace{-0.15cm}
The UM can be used either directly as an approximation of probabilities or it can be used as a proposal for amortised inference. This abstract proposes an idea of automatic generation and training of a neural network given a probabilistic program and samples from its prior, such that {\bf one} neural network can be used as a proposal for performing the posterior inference given any possible evidence set. Such framework could be implemented in one of probabilistic programming platforms, e.g. in Pyro. While this approach directly could be applied only to the models with bounded number of random choices, it might be possible to map the names of random choices in a program with finite but unbounded number of those random choices to the bounded number of names using some schedule, hence performing a version of approximate inference in sequence.

\label{sec:cite}

\bibliography{refs}

\newpage
\appendix
\section{Appendix}
\label{appendix:code}
\subsection{Implementation}

\begin{figure}[ht]
        \centering
        \begin{lstlisting}[language=Python]
        def probProg(t1, v):
           for i in [2, 3, ..., 50]:
                if abs(t[i-1]) < 1:
                    t[i] ~ Bernoulli(abs(t[i-1]))
                else:
                    t[i] ~ Gaussian(t[i-1],v)
            return t1, t2, ..., t50
        \end{lstlisting}
        \caption{Example for probabilistic program.}
\end{figure}

\begin{figure}[ht]
        \centering
        \includegraphics[width=120mm]{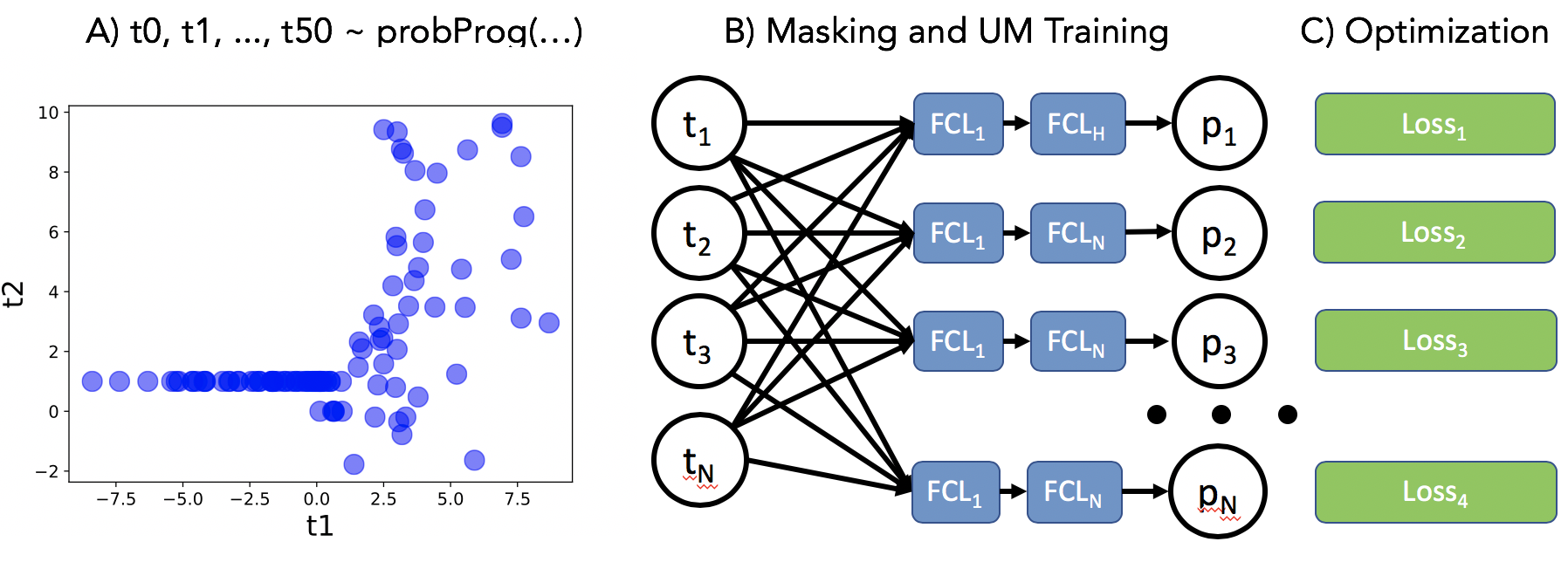}
        \caption{UM learning algorithm. The samples from the program are masked and used to train the UM with different losses for each output. Note that only the first two components are shown in the scatter plot.}
        \label{example_pp}
\end{figure}

\begin{figure*}[ht]
\footnotesize
\begin{lstlisting}[language=Python]
def probProg(t0, v):
  for i in [2, 3, ..., 50]:
    if abs(t[i-1]) < 1:
      t[i] ~ Bernoulli(abs(t[i-1]))
    else:
      t[i] ~ Gaussian(t[i-1],v)
   return t0, t1, ..., t50

def inputGen():
   t0 ~ Gaussian(0, 3)
   v ~ Gamma(3, 1)
   return t0, v

UM = generateUM(probProg)
UM.train(probProg, inputGen)
\end{lstlisting}
\caption{Building probabilistic program and UM training.}
\label{fig:buildUM}
\hfill
\begin{lstlisting}[language=Python]
# UM for approximating P(Y|X) 
# Evidence (except t5 and t8):
data = {t1=0.3, ... , t50=0.5}

y5, y8 = UM.condMarginals(
  observ = data,
  sites = ["t5", "t8"]
)

# UM + Importance Sampling
y5, y8 = Sample(
  model = probProg,
  guide = UM.guide,
  observ = data,
  sites = ["t5", "t8"],
  num_samples=100000
)
\end{lstlisting}
\caption{Conditional posteriors and guide distribution.}
\label{fig:applyUM}
\end{figure*}

\end{document}